# Multi-Stakeholder Disaster Insights from Social Media Using Large Language Models

Loris Belcastro, Cristian Cosentino, Fabrizio Marozzo*, Merve Gündüz-Cüre, Sule Öztürk-Birim

*Abstract*—In recent years, social media has emerged as a primary channel for users to promptly share feedback and issues during disasters and emergencies, playing a key role in crisis management. While significant progress has been made in collecting and analyzing social media content, there remains a pressing need to enhance the automation, aggregation, and customization of this data to deliver actionable insights tailored to diverse stakeholders, including the press, police, EMS, firefighters, and other decision-makers. This effort is essential for improving the coordination of activities such as relief efforts, resource distribution, and media communication. This paper presents a methodology that leverages the capabilities of Large Language Models (LLMs) to enhance disaster response and management. Our approach combines classification techniques with generative AI to bridge the gap between raw user feedback and stakeholder-specific reports. Social media posts shared during catastrophic events, including earthquakes, floods, fires, and hurricanes, are analyzed with a focus on user-reported issues, service interruptions, and encountered challenges. We employ full-spectrum LLMs, using analytical models like BERT for precise, multi-dimensional classification of content type, sentiment, emotion, geolocation, and topic. Generative models such as ChatGPT are then used to produce human-readable, informative reports tailored to distinct audiences, synthesizing insights derived from detailed classifications. We compare standard approaches, which analyze posts directly using prompts in ChatGPT, to our advanced method, which incorporates multi-dimensional classification, sub-event selection, and tailored report generation. Our methodology demonstrates superior performance in both quantitative metrics, such as text coherence scores and latent representations, and qualitative assessments by automated tools and field experts, delivering precise, impactful insights for diverse disaster response stakeholders.

## I. Introduction

Social media analysis has become a cornerstone for understanding societal dynamics and user behavior. Each post shared on these platforms contains valuable information, such as the topics discussed, the sentiments conveyed, and the challenges highlighted [1]. Consequently, both researchers and industry practitioners have invested heavily in advanced machine learning techniques to extract actionable insights from this data. Social media, in particular, has emerged as an indispensable real-time information source for managing complex scenarios, including emergency situations, natural disasters, and catastrophic events [2], [3]. Furthermore, it plays a pivotal role in disaster preparedness within smart cities by enhancing early warning systems, optimizing resource allocation, and supporting predictive modeling to mitigate risks. However, despite significant advancements in the classification and aggregation of social media content, critical gaps remain in automating, organizing, and tailoring citizen-reported issues for specific needs. Addressing these gaps can significantly improve the efficiency of disaster response efforts, including rescue operations, resource allocation, and multi-stakeholder communication [4].

Large Language Models (LLMs) play a strategic role in analyzing text by enabling the classification, aggregation, and enrichment of information in user posts [5]. Analytical models, such as BERT (encoder-based), excel at understanding linguistic context, allowing for precise classification and categorization across multiple dimensions. On the other hand, generative models, such as ChatGPT (decoder-based), serve as powerful text generators that enhance data presentation and interpretation, improving information synthesis and facilitating rapid decision-making. Additionally, the real-time processing capabilities of these models are crucial for the early detection of sub-events and the automatic generation of detailed reports [6], which are indispensable during crises.

In this paper, we propose a novel methodology to leverage LLMs for analyzing social media posts during and after catastrophic events, with the goal of generating comprehensive, stakeholder-specific reports. Our approach combines the strengths of encoder-based models for precise, multi-dimensional classification with decoder-based models for generating structured, human-readable reports. The methodology begins by collecting social media posts relevant to disaster-affected areas, capturing user-reported issues such as collapsed buildings, damaged infrastructure, and service outages. These posts are processed using BERT models to classify content across multiple dimensions, including post type, geolocation (explicit or implicit), sentiment, and key topics. This step provides a detailed understanding of the problems reported by citizens. Subsequently, a generative LLM, such as ChatGPT, synthesizes these insights into comprehensive, actionable reports tailored to diverse stakeholders, such as press outlets, emergency responders, and government agencies. Additionally, a chatbot interface enables stakeholders to analyze the posts and the generated reports interactively through a question-and-answer approach, facilitating deeper insights and tailored exploration of the content.

L. Belcastro, C. Cosentino, and F. Marozzo are with the Department of Informatics, Modeling, Electronics, and Systems Engineering (DIMES), University of Calabria, Rende (CS) 87036, Italy. (Email: loris.belcastro@unical.it, cristian.cosentino@unical.it, fabrizio.marozzo@unical.it).

M. Gündüz-Cüre and S. Öztürk-Birim are with Manisa Celal Bayar University, Manisa 45040, Turkey. (Email: merve.gunduz-cure@mcbu.edu.tr, sule.ozturk-birim@mcbu.edu.tr).

*Corresponding author.

Extensive experiments on various datasets have demonstrated the effectiveness of our methodology in detecting events and issues following a disaster and generating detailed, immediate reports. Unlike basic reports generated by systems like ChatGPT, our approach enriches the preliminary classification of posts across multiple dimensions, enabling the generative LLM to produce more precise and insightful reports. Evaluated using a comprehensive set of metrics, our methodology outperforms reports generated by ChatGPT from raw posts, excelling in quantitative measures such as text scores and latent representations, as well as qualitative assessments by automated tools and field experts.

In comparison with the state of the art, our approach goes beyond existing single-step classification or basic prompt-based summarization tools by combining classification and generative large language models to deliver real-time, context-rich reports. We integrate multidimensional data enrichment, including sentiment analysis, topic modeling, and NER, to enable granular event localization and actionable insights for diverse stakeholders. Evaluations on multiple crisis datasets demonstrate the superiority of our methodology in terms of accuracy, clarity, and operational relevance, fostering effective decision-making and improved collaboration during large-scale emergencies.

The paper is organized as follows. Section II discusses related work and highlights the differences between our methodology and existing research. Section III outlines the proposed methodology. Section IV presents the results, while Section V provides the conclusion.

## II. Related work

Many recent studies have highlighted how social media is an important tool for organizing rescue operations when managing natural disasters and catastrophic events. In fact, social media allows for the sharing of detailed information in real time and on a large scale, which is essential for the timely resolution of critical situations [7]–[9]. However, using social media data is a complex process with several pitfalls. The huge volume of data and the speed at which it is generated make the collection and analysis phases challenging. Additionally, the data collected may not be immediately ready for the analysis, but requires appropriate methods to effectively select, transform, enrich, and organize it.

Numerous studies have recently focused on leveraging social media to improve the efficiency of emergency response operations. These studies analyze the main challenges associated with using social media data in disaster contexts, including the complexity of processing large amounts of data in a timely manner, the presence of unwanted or false information, and the difficulties in collecting data that document the various phases of a disaster [10]–[13].

Further investigations explored the complexities related to the analysis of social media posts during large-scale emergencies, focusing on various aspects such as the coordinated management of evacuation operations [14], the integration of data from different sources [15], and the analysis of the dynamics of information diffusion during such events [16].

In particular, natural disasters such as earthquakes have attracted significant attention among the critical situations addressed by studies related to disaster management, with the proposal of numerous systems that exploit collaborative and information-sharing tools, such as social media and user feedback, to detect seismic events and evaluate their impact on people and infrastructures [17], [18].

Other studies have focused on predicting and managing urban floods [19], [20], which can be caused by various factors, including heavy rainfall, ineffective stormwater drainage, malfunctioning drainage infrastructure, or natural events such as sea level rise.

In disaster management, it is often useful to identify so-called *sub-events*, which refer to specific and localized events (e.g., a building collapse, a gas pipeline explosion, power outages) that occur within a larger critical event (e.g., an earthquake or a hurricane) [21]. Several studies have focused on detecting sub-events from social media data, using both supervised and unsupervised methods. Supervised methods typically employ weighted graphs [22] and neural networks to identify, categorize, and summarize sub-events in social media content [23]–[25]. Additionally, other approaches incorporate semantic modeling and event-specific feature extraction to enhance detection accuracy and contextual relevance [26]. Although these methods can be effective, they often require significant effort in terms of configuration and optimization, which can hinder their overall effectiveness. Consequently, many researchers have focused on unsupervised methods for sub-event detection. In this context, clustering algorithms are commonly used to analyze textual and geolocation data from social media [25], [27]. Other approaches involve topic modeling, employing established algorithms like LDA (Latent Dirichlet Allocation) and HDP (Hierarchical Dirichlet Processes) to extract sub-events by analyzing the semantic content of documents [28].

A key challenge in leveraging social media posts for disaster management is extracting the location of events to enable targeted interventions [29]. Social media data often lack accurate geolocation information, with geotags frequently missing or incorrect, and the user's current location not necessarily aligning with that of the disaster [30], [31]. Consequently, it is crucial to extract location mentions directly from the content of social media messages. Traditional approaches have used pre-trained named entity recognition (NER) tools like Stanford NER[1] and SpaCy NER[2], as well as deep learning models, to identify geographic references [32], [33]. Advances in transformer-based models, such as BERT, have further enhanced geolocation accuracy in social media content [8], [34]. Additionally, recent research has explored integrating geoknowledge with LLMs like ChatGPT to improve location extraction from disaster-related posts [35], demonstrating the potential of these tools to compensate for the lack of geotagging and metadata.

Recent advancements have emphasized the need for integrating social media data into disaster management systems,

---

[1] https://nlp.stanford.edu/software/CRF-NER.html
[2] https://spacy.io

particularly for identifying and categorizing first responders such as police, EMS, and firefighters [36]. Systems like FReCS (First Responder Classification System) have demonstrated the potential of using BERT-based models to classify disaster-related tweets and assign roles to first responders with enhanced precision and scalability [36]. These classifications enable the creation of tailored strategies for resource deployment and improve coordination among diverse stakeholders. Such frameworks address challenges like the absence of structured classifications and the complexity of coordinating large-scale emergency responses, as highlighted during events like Hurricane Harvey and Hurricane Maria [7], [37].

Compared to our previous work [6], which primarily focused on using generative LLMs for disaster management, this study introduces significant advancements. Specifically, it expands the scope of analyzing various disaster scenarios by strategically integrating encoder-based (BERT-like) models and decoder-based generative models (ChatGPT/GPT-4) in a complementary manner. Furthermore, we systematically evaluate the generated reports, now enriched with contextual data such as external resources, sub-event details, and dynamic user prompts, and compare them against conventional methods. The new approach yields superior accuracy in summarizing events, highlighting critical issues, and delivering context-aware insights from social media data to multiple stakeholders through advanced large language models. The key contributions of this research are:

- We present a comprehensive methodology for processing and analyzing social media data related to different disaster events, enabling the generation of reliable reports that accurately describe sub-events.
- We propose a multidimensional data enrichment approach that enhances generative AI models' ability to produce detailed reports by integrating sentiment analysis, topic modeling, and named entity recognition (NER) for improved event localization and identification.
- We generate structured and automated reports that effectively summarize key details of natural disasters, providing clear and concise insights into critical aspects such as event timelines, affected areas, severity levels, and sub-events.
- We introduce a stakeholder-centric design that tailors the final reports to the specific needs of various decision-makers, including press outlets, emergency services, and other agencies, ensuring that the content is highly relevant and actionable in diverse operational contexts.
- We demonstrate that our approach outperforms traditional methods by using a combination of quantitative metrics, such as text coherence scores and latent representations, alongside qualitative assessments from automated tools and field experts, resulting in more accurate and reliable reports.

## III. Methodology

The methodology proposed in this work leverages issues reported by users in social media posts to enhance disaster management and enable targeted interventions. Specifically, feedback from users following a catastrophic event are used to create a detailed report that outlines key issues and critical situations, thereby supporting a user-centric and data-driven approach. The execution flow of this methodology, illustrated in Figure 1, is organized into three main phases: ($i$) monitoring a disaster event by collecting posts from social media platforms; ($ii$) employing analytical LLMs for multi-dimensional classification to identify citizen-reported issues; and ($iii$) exploiting generative LLMs to synthesize the classified data into informative and actionable reports tailored to various stakeholders.

The first phase involves systematically collecting relevant social media posts related to a disaster, focusing on posts generated in the affected area. The collection of posts is carried out on social media platforms employing keywords or geographical metadata associated with the disaster. To ensure the appropriateness of the dataset for the subsequent phases, we apply a filtering mechanism to select only the relevant posts, such as those from users residing in the affected region. This data collection phase is crucial for obtaining accurate results in the subsequent phases of classification, localization, and reporting.

Next, the second phase employs multi-dimensional classification [38] to identify citizen-reported issues. In this phase, we utilize Encoder-Based Analytical Models (e.g., BERT), which are particularly effective for tasks such as classification, sentiment analysis, and information retrieval. These models excel at transforming textual data into dense vector representations that capture contextual meaning, enabling a comprehensive analysis and reliable categorization of the posts. Specifically, BERT models are used to classify posts across multiple dimensions: content type (e.g., news or opinion), sentiment (e.g., positive or negative), emotion (e.g., joy, anger), catastrophic event (e.g., infrastructure damage), and location through Named Entity Recognition (NER). To achieve accurate classifications, we employ fine-tuned BERT models tailored to each specific dimension. Additionally, BERTopic is utilized to identify the topics discussed in the posts, offering a high-level understanding of thematic trends and emerging issues. This thorough classification process provides a detailed understanding of the issues, enabling the identification of sub-events and their global descriptions by grouping similar events or those occurring in the same locations.

Finally, the third phase of our methodology focuses on dynamically generating detailed and customized reports using generative models like GPT-4. By leveraging these models via their APIs, we design tailored prompts to produce reports that meet the unique needs of diverse stakeholders, including media operators, police, EMS, firefighters, and other operational groups. To generate these customized reports, classified posts are carefully selected based on the report's scope and enriched with relevant external information, such as meeting points, escape routes, emergency service contact details, and historical data. This integration of context enhances the quality, relevance, and overall utility of the final reports. Additionally, a chatbot interface allows operators to refine reports and retrieve additional insights through interactive question-and-answer sessions. This iterative approach not only enhances the understanding of the generated reports but also adapts their content to evolving needs, improving situational awareness and supporting informed decision-making during crisis response. For example, EMS teams may request concise summaries of incidents within a specific location, while firefighters may require detailed information about nearby water sources, building layouts, and access routes to better coordinate their operations. To implement this system efficiently, strategies such as Retrieval-Augmented Generation (RAG) can be considered depending on the size and structure of the dataset. For small, well-organized datasets, embedding the entire dataset directly into the input prompt may be sufficient, as modern LLMs like GPT-4 can process thousands of tokens in a single interaction. However, as data grows, RAG becomes fundamental for optimizing token usage, retrieving only the most relevant data items, and improving query response accuracy to maintain both efficiency and relevance. By choosing the most appropriate strategy, the system can effectively balance efficiency, relevance, and scalability to meet the demands of real-time crisis management.

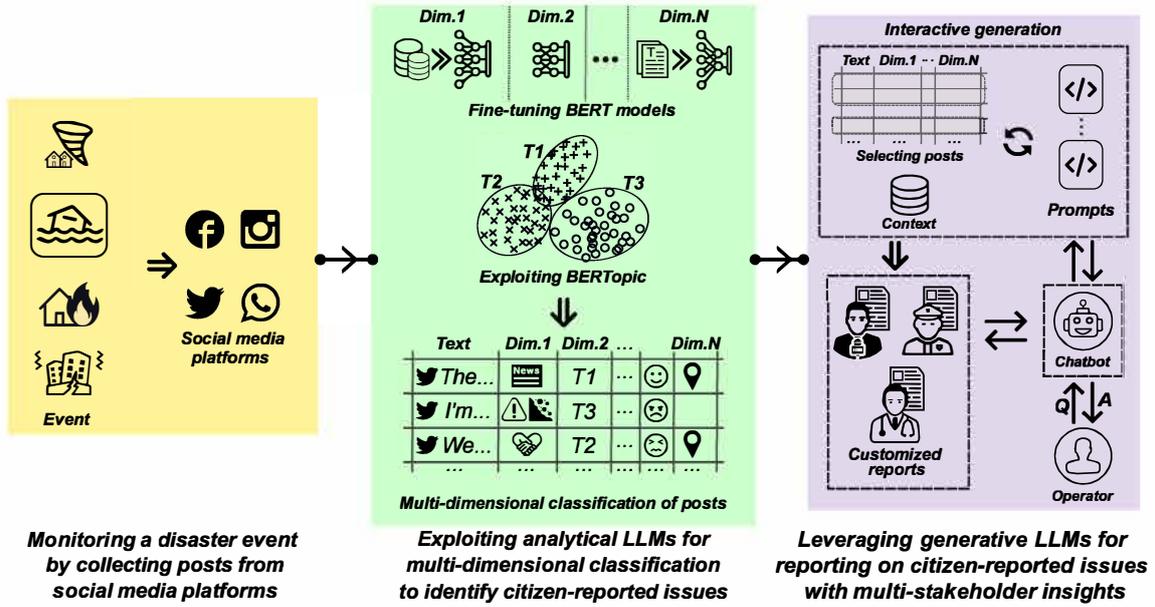

Fig. 1. Execution flow of the proposed methodology.

## IV. EXPERIMENTAL RESULTS

The experimental evaluation of the proposed methodology is structured as follows. Section IV-A presents the dataset used in our experiments and describes the pre-processing steps taken. Next, Section IV-B discusses the multi-dimensional classification process employed to enrich the original dataset. This process leverages BERT-based language models to classify posts according to various aspects such as content type, sentiment, emotion, sub-event detection and topic analysis IV-B2). Section IV-C describes how to generate a detailed report starting from posts written by users, appropriately classified, about a disastrous event.

### A. Data description

In natural disaster research, several datasets containing social media posts published by users immediately after catastrophic events have been compiled and made available over the years. For some events, like hurricanes, which are predictable in advance, monitoring campaigns can be planned using data collection processes on social media platforms with specific keywords or hashtags. In contrast, unpredictable events like earthquakes are often monitored using systems that continuously track general keywords or hashtags (e.g., *hurricane* or *#hurricane*). To validate our methodology, we utilized the HumAID (Human-Annotated Disaster Incidents Data) [39] dataset, a well-known archive consisting of approximately 77,000 manually labeled tweets drawn from a larger pool of over 24 million tweets. Such tweets were collected during 19 major real-world disasters from 2016 to 2019, including hurricanes, earthquakes, fires, and floods. This extended classified dataset is exciting because it likely includes posts from users in affected areas during different disasters, providing valuable firsthand information from eyewitnesses or those directly involved in critical situations [40].

The tweets included in the HumAID dataset are classified using the several categories related to natural disasters, including: $i$) *caution and advice*, which encompasses notices that are issued or revoked; $ii$) *sympathy and support*, identifying tweets offering prayers, thoughts, and emotional support; $iii$) *requests or urgent needs*, which includes reports of urgent necessities such as food, water, clothing, money, medicine, or blood; $iv$) *infrastructure and utility damage*, which identifies damages to buildings, roads, bridges, power lines, communication poles, or vehicles; $v$) *rescue volunteering or donation effort*, capturing tweets about rescue efforts, volunteering, donation activities, safe transport, evacuation, medical and food assistance, and more; $vi$) *not humanitarian*, which includes tweets that do not convey information related to humanitarian aid; $vii$) *displaced people and evacuations*, covering posts discussing situations where people have had to change residence due to the crisis (e.g, evacuations); *viii*) *injured or dead people*, which reports of people who are injured or have died as a result of the disaster; and *ix*) *missing or found people*, reporting individuals who are missing or have been found.

It is worth noting that the HumAID dataset includes two other classes, namely *don't know can't judge* and *other relevant information*, which we excluded from our analysis as they were considered outliers.

*B. Multi-Dimensional Classification Using BERT Models*

As discussed in Section III, we use multidimensional classifiers to enrich social media posts with categories derived from the text. These categories help identify issues and sub-events related to catastrophic events, enhancing the ability of LLMs to produce accurate and relevant reports. In addition to text-derived dimensions, we incorporate metadata such as timestamps, user information, and engagement metrics (likes, reposts, and favorites). Together, these dimensions enable more precise data filtering and improve the relevance of the insights generated. Specifically, we train and utilize classifiers for the following text-derived dimensions:

- *Content Type*, which distinguishes between factual content, such as news, and subjective opinions expressed by users (classes: *news* or *opinion*).
- *Sentiment*, determining whether a post conveys a positive or negative sentiment (classes: *positive*, *negative*).
- *Emotion*, which identifies the emotional tone and expressions conveyed within the text (classes: *anger*, *anticipation*, *disgust*, *fear*, *joy*, *sadness*, *surprise*, *trust*).
- *Disaster Event*, which classifies posts into the nine specific disaster-related HumAID classes, such as *caution and advice*, *sympathy and support*, and *infrastructure and utility damage*, which have been discussed in the last paragraph of the previous section. Each class represents different types of information, feedback, or issues related to a disaster.
- *Sub-Event*, a binary classification that distinguishes between posts indicating sub-events (happened during or immediately after a disaster) and those that do not. To train this classification model, we started from the HumAID dataset and created a new labeled dataset as follows: posts labeled as *infrastructure and utility damage*, *displaced people and evacuations*, *injured or dead people*, or *missing or found people* in the HumAID dataset have been labeled as *sub-event post*; the others were labeled as belonging to the second category.
- *Named Entity Recognition (NER)*, identifying and categorizing named entities in text, such as people, organizations, and locations (e.g., states, regions, cities, streets). NER also aims to reconstruct detailed or partial location information by contextualizing and enriching mentions of places. For instance, if a text refers to a specific square, street, or establishment, NER can deduce the most complete address possible by integrating additional geographic context, such as neighborhood, city, state, and country. For example, encountering *Caffè Strada* in a post about an event in California may allow reconstruction as *Caffè Strada, 2300 College Ave, Berkeley, CA, United States*. If full details are uncertain, a partial reconstruction, such as *Elmwood, Berkeley, CA, United States* where *Elmwood* is a district of Berkeley, is provided, ensuring accuracy and avoiding over-speculation.
- *Stakeholder Identification*, which focuses on detecting the various parties involved in or affected by the events described. This classification is derived from the FReCS dataset[3], which provides a rich taxonomy of stakeholder types and subtypes. We consolidate these into five primary categories, including *Police*, *EMS*, *Firefighter*, *Media*, and *Government\Organization*. Such classification supports the tailoring of responses, ensuring that each identified stakeholder group receives information and resources best aligned with their objectives, capabilities, and responsibilities.
- *Topic*, which identifies and associates the subject matter discussed in a post. In this context, predefined classes cannot be established, necessitating a dedicated topic extraction process. To achieve this, we exploit BERTopic, a state-of-the-art topic modeling approach that uses embeddings and clustering to discover coherent and meaningful topics from textual data. This process has been extensively discussed in Section IV-B2.

To ensure accurate categorization, for each text-derived dimension, we chose the most accurate classifiers in the literature. In particular, as recommended in [41], we leveraged BERTopic for topic extraction. BERTopic is a modern topic modeling technique that leverages transformer-based language models, such as BERT or RoBERTa, to generate dense document embedding capable of capturing the semantic content of the documents and to understand relationships between words and phrases. To extract the *NER*, we used bert-large-NER [42], a fine-tuned BERT model that achieves state-of-the-art performance in the named entity recognition task. This model has been developed on the CoNLL-2003 dataset, used to identify entities such as location (LOC), organization (ORG), person (PER) and miscellaneous (MISC). For the *topic* and *NER* dimensions, we employed pre-existing models for inference, whereas additional fine-tuning was necessary for the other dimensions to adapt the models to specific tasks. More details about this fine-tuning process are discussed in the next section.

*1) Fine-Tuning BERT Models:* For dimensions other than *NER* and *topic*, we used BERT-based models [43] for classifying social media posts, as they have proven effective in capturing the semantic and syntactic features of microblog texts. For each dimension, a fine-tuning process was performed to adapt the models to specific tasks. Additionally, an in-depth comparison of different BERT-based models was carried out to choose the one providing the best performance.

For each dimension a specific dataset was used for fine tuning. Specifically, for the *content type* dimension, we refined the BERT models using a publicly available dataset of 30,000 tweets[4], classified as either informational content (e.g., news) or personal opinions. For the *sentiment* dimension, we used another dataset of 16,000 tweets, labeled as positive or negative[5]. For the *disaster event* dimension, we employed a dataset consisting of 10,800 tweets classified into nine different categories, derived from the HumAID [40] dataset. In this case, since the original dataset was unbalanced, we undersampled the original dataset to create a balanced dataset containing 1,200 instances for each class. Finally, for the *sub-event* dimension, we trained the models using the HumAID dataset reconfigured into a two-class dataset, as described in Section IV-A.

---

[3]https://github.com/abdul0366/FReCS/tree/main/Dataset

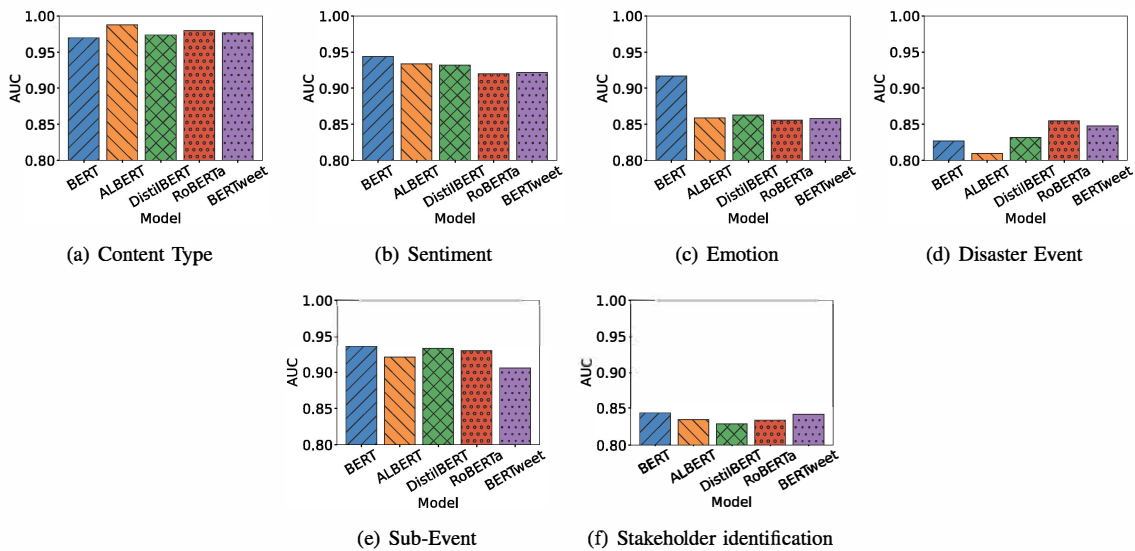

Fig. 2. Comparison of BERT-based models in terms of AUC across different dimensions.

Figure 2 illustrates the performance comparison among different BERT-based models, specifically BERT, DistilBERT, RoBERTa, ALBERT, and BERTweet, evaluated in terms of Area Under the Curve (AUC). Each model exhibits slight variations in performance depending on the classification task, attributed to their distinct architectures and training methodologies. Notably, ALBERT proved to be the most effective model in classifying *content type*, though its superiority was marginal. In contrast, BERT excelled in *sentiment*, *emotion*, *sub-event*, and *stakeholder* classification tasks, outperforming the other models in these areas. Furthermore, RoBERTa achieved the highest AUC for the *disaster event* classification task, which involves nine classes.

*2) Performing Topic Analysis:* In the present study, we use BERTopic to identify topics in social media data related to disasters. An important decision in topic modeling is determining the optimal number of topics for all datasets. This decision is a compromise between producing overly general categories with a smaller number of topics and constructing overlapping or similar categories with a larger number of topics. To determine the optimal number of topics, we evaluated different metrics, including coherence. The coherence value (CV) helps distinguish separate topics from one another by reflecting the coherence within topics. In particular, high values of CV typically indicate that the topics are more meaningful and interpretable.

Figures 3 illustrate the CVs for different types of disasters when increasing the number of considered topics. As shown, around 25-30 topics yield the highest coherence values. For each disaster, we selected the number of topics ensuring the highest CV value, since this number provides specific topics that appropriately capture disaster details in social media posts with greater granularity than fewer topics.

From the analysis, it emerged that there are both common topics shared across multiple datasets and unique ones specific to each disaster. The most frequently observed category, common to all datasets, is the demand for relief, prayers, and help for the victims of the disasters. Another notable category pertains to the death toll, providing information about the number of people affected and killed, especially in wildfires and earthquakes, which were more destructive in terms of human lives compared to other disasters. In contrast, after hurricanes and floods, people discussed damaged areas and infrastructures (e.g., houses, bridges, and roads) more frequently.

Another commonly observed topic is the discussion related to aid coming from foreign countries. This topic was particularly noted for disasters outside the US. During the earthquake in Mexico and the floods in Kerala and Sri Lanka, people focused on the help from various countries and related issues. Conversely, during and after the California wildfire, discussions often criticized improper usage of troops for disaster management. Another interesting category that was commonly observed involved celebrity relief efforts. For all the disasters, donations or campaigns by various celebrities were frequently discussed as prominent topics. In all cases, people expressed their gratitude to those who provided help.

[4]https://www.kaggle.com/datasets/ferno2/training
[5]https://www.kaggle.com/datasets/vinayakshanawad/us-news-dataset

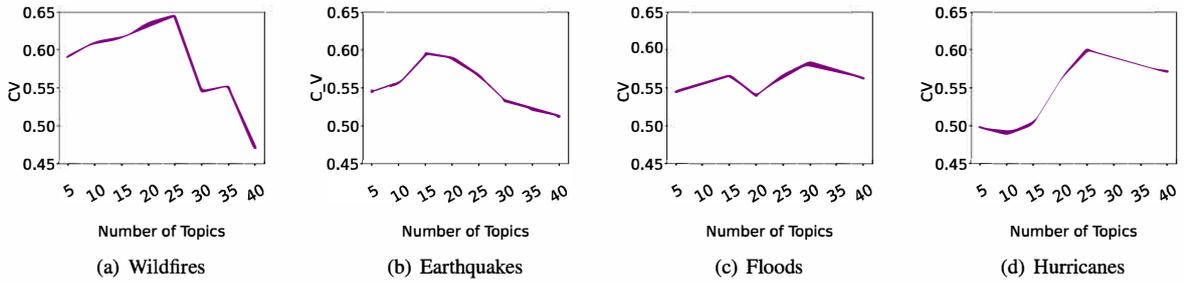

Fig. 3. Coherence Value (CV) in topic modeling with a variable number of topics.

*3) Social media data classification:* Table I presents five examples of tweets that have been classified on different dimensions. In particular, for each tweet, top three representative words (or phrases) of the topic identified in the text, type of content (news or opinion), sentiment (positive or negative) and emotion, class of disaster event and whether it refers to a sub-event or not, and location details provided by NER are reported.

The first and second tweets discuss the California fires with different focus and tone, targeting media stakeholders. In particular, the former expresses frustration and criticism regarding the use of troops on the border instead of assisting in the fight against wildfires in California. The second expresses disbelief at the rapid spread of the fire in California and offers prayers for people affected. The third tweet, relevant to police and government organizations, describes the long lines for relief efforts outside the McMahon Stadium in Calgary (Canada). It is categorized as news with negative sentiment and an emotion of anticipation. The disaster caused damages to infrastructures and services. The fourth tweet, targeting firefighters and emergency medical services (EMS), reports the rescue of a little girl from the rubble of the earthquake in Italy. It is classified as news and expresses joy. The disaster is represented by people missing or found, with Italy identified as a named entity location. Finally, the last example, relevant to government organizations, announces the opening of the Burger Stadium in Austin (USA) as a center for displaced people with urgent need for supplies. The disaster event is characterized by displaced people and evacuations, with the Burger Stadium in Austin highlighted.

### C. Disaster Reporting Using Generative Models

After classifying and enriching social media data with various useful information (e.g., the nature of the disaster, occurred sub-events, locations, and covered topics), the proposed methodology can generate dynamic and interactive reports tailored to the needs of different stakeholders. These reports are designed to support decision-making processes by providing actionable insights that enable the formulation of effective response strategies. The report generation process leverages generative LLMs for creating coherent and comprehensive texts by aggregating and synthesizing diverse information. Reports can be generated using one or more prompts, each contributing to different sections or aspects of the report.

Among the models provided by OpenAI, we chose GPT-4o for its advanced capabilities, flexibility, and widespread use.

We employ two primary approaches in our methodology, the *basic approach* and the *advanced approach*, with details of each outlined below.

*a) Basic Approach (ChatGPT-4o + File with all posts):* This standard approach utilizes the interactive capabilities of GPT-4o (ChatGPT-4o). The input consists of a comprehensive file containing all posts related to the disaster event, combined with carefully crafted prompts designed to analyze or describe specific aspects of these posts. This method is highly accessible, allowing users to input the list of posts and employ prompt-based queries to extract meaningful insights.

In this approach, ChatGPT is tasked with acting as a *report writer* to generate various types of reports. The following are two examples of reports that can be created using this method.

The first report is designed for *media operators* and focuses on two primary objectives: describing the key topics of discussion and analyzing user opinions, particularly in terms of sentiment and emotions expressed. The report begins with an introductory section that provides a comprehensive overview of the social media posts, highlighting the main *topics* discussed by users. It then goes into an in-depth analysis of the *opinions* (sentiment and emotions) expressed, presenting both the negative and positive perspectives on the event and exploring the emotions conveyed in the posts.

To achieve this, two specific prompts are used: $i$) one to identify the topics of discussion (`topic_prompt`); and $ii$) another to assess the sentiment and emotional responses of users (`opinion_prompt`). These prompts take as input the disaster event ($E$), the affected area ($A$), and the date range ($D$) during which it occurred. The prompts are detailed below:

> **`topic_prompt`**: "Produce an analytical report summarizing the key topics discussed in posts by citizens living in the affected area ($A) during the disaster event ($E) within the date range ($D). The report has to be constrained to ($W) words, presented in paragraph format without subsections. Exclude a conclusion section. Use the attached file containing social media posts as input."
> **`opinion_prompt`**: "Generate an analytical report detailing the sentiments and emotions expressed by users regarding the disaster event ($E) in the affected area ($A) within the date range ($D). The report has to be

TABLE I
EXAMPLES OF POST CLASSIFICATIONS ACROSS THE ANALYZED DIMENSIONS.

| Post | Topic (Top-3 words) | Content Type | Sentiment & Emotion | Disaster Event / Sub-Event | NER Location | Stake-holder |
|---|---|---|---|---|---|---|
| *In all seriousness, why the hell is the military still at the border and not in California helping to fight the fires?* | wildfire, california, homes | Opinion | Negative Anger | Not humanitarian | California USA | Media |
| *My heart really goes out to the victims of the Camp Fire in California who have lost so, so much. So sorry for all you are going through, please stay safe.* | wildfire, california, homes | Opinion | Positive Optimism | Sympathy and support | California USA | Media |
| *Long lines expected for fire relief cards as hundreds of people will wait outside McMahon's Stadium in Calgary* | wildfire, mcmahon, support | News | Negative Anticipation | infrastructure and utility damage (sub-event) | McMahon Stad. Calgary Canada | Police Gov.\Org. |
| *#BREAKING: Italian firefighter chief said a 8 y/o girl was pulled out alive from the rubble. #ItalyEarthquake* | earthquake, victims, death toll | News | Positive Joy | missing or found people (sub-event) | Italy | Firefighter EMS |
| *Burger Stadium in Austin now open as evacuee center. Urgent need: diapers, sanitary products, clothing, shoes. No food. Drop off at entrance* | hurricane, victims, shelter | News | Positive Optimism | displaced people and evacuations (sub-event) | Burger Stad. Austin USA | Gov.\Org. |

constrained to (*$W*) words, presented in paragraph format without subsections. Exclude a conclusion section. Use the attached file containing social media posts as input."

The second report is designed for *emergency services* and concentrates on analyzing sub-events in a specific city $C$ affected by the disaster $E$. This analysis aims to provide actionable insights tailored for emergency responders such as the police, EMS, and firefighters. The prompt used for this report is presented below:

`"city_subevent_prompt"`: "Produce an analytical report to support three key stakeholders — Police, EMS, and Firefighters — by summarizing actions taken, issues reported by citizens, and tasks to address in the affected location (*$C*) during the disaster event (*$E*) within the data range (*$D*). The report has to be constrained to (*$W*) words, presented in paragraph format without subsections. Exclude a conclusion section. Use the attached file containing social media posts as input. Insert in the report references to original posts using a bibliographic style (e.g., [1] $post_1$, [2] $post_2$)."

In this approach, ChatGPT is tasked with understanding user queries, analyzing provided files, identifying the most relevant posts, and generating comprehensive reports based on defined prompts. When using ChatGPT-4o with file inputs, certain limitations should be noted. For instance, the model cannot directly process complex textual file formats such as deeply nested JSON structures, intricate XML schemas, or Markdown files with extensive formatting and links. To make such content usable, it needs to be preprocessed and converted into simpler formats, such as plain text or structured data formats (e.g., CSV or TSV), compatible with prompt requirements. Similarly, large spreadsheets or extensive sets of tabular data require appropriate operations to identify relevant regions, simplify the structure, and handle formatting issues, such as cell merging or data types, to ensure compatibility with LLMs [44]. Additionally, the context length of the model imposes restrictions on handling large datasets or extensive file content, as the token limit determines the amount of information it can process at once. Even within the token limit, performance may decline when the token count approaches the maximum, especially for tasks involving long and complex content [45]. While file uploads of up to 512 MB are allowed, ChatGPT-4o can only process the portion of the file that fits within its 128,000-token context window (approximately 96,000 words or 6,400 tweets, assuming 15 words per tweet). Content exceeding this limit remains unprocessed. In this study, the datasets fit comfortably within the context window, enabling the system to accommodate the entire set of tweets.

*b) Advanced Approach (GPT-4o with API + Most representative posts):* This approach utilizes GPT-4o with in-context learning via the API, where the token limit is 128,000 tokens for the context length and 16,384 tokens for the output[6]. To work within these limits, each prompt is paired with a filtered and relevant data sample, carefully tailored to the analysis. These samples include:

- *Sub-events in a specific city*: Posts are filtered to include those classified as sub-events within the input sample and mentioning the city under analysis (e.g., using NER information).
- *Emotions*: Posts are filtered to include those classified as expressing a specific emotion, such as anger, which may arise from a perceived lack of intervention or inadequate prevention measures. The emotional dimension is categorized into eight distinct classes: anger, anticipation, disgust, fear, joy, sadness, surprise, and trust.
- *Sentiment analysis*: Posts are analyzed based on the sentiment dimension, which includes two primary classes: positive and negative. To ensure consistency, the sample distribution mirrors that of the original dataset. For example, if the dataset consists of 75% negative posts and 5% positive posts (other neutral), the sample preserves this proportion.

---
[6]https://platform.openai.com/docs/models/gpt-4o

Similar filtering methods can be applied to other dimensions, such as topic or location, ensuring that the data is aligned with the intent of the prompt.

To calculate the most relevant posts for analysis among the dimensions deemed of interest for analysis, we use the following method. Consider an initial dataset of posts $D$, where each post $p \in D$ is associated with one or more dimensions $d_1, d_2, \ldots, d_k$. Each dimension $d_i$ has a set of possible classes $C(d_i) = \{c_1, c_2, \ldots, c_m\}$. For each post $p$ and each dimension $d_i$, there is an associated probability distribution over the classes $c \in C(d_i)$, denoted as $P(c \mid d_i)$.

To create a representative sample $S$ of $N$ posts for analysis:

1) *Select dimensions and classes:* The user identifies one or more dimensions of interest $\{d_1, \ldots, d_z \mid z \geq 1\}$ (not necessarily all), which are considered relevant for analysis. For each selected dimension, specific classes $C'(d_i) \subseteq C(d_i)$ may also be chosen based on the scope of the analysis.
2) *Compute class distributions:* For each selected dimension $d_i$, calculate the probability $P(c \mid d_i)$ of posts in $D$ that belong to each class $c \in C'(d_i)$.
3) *Allocate sample sizes:* For each class $c \in C'(d_i)$, determine the number of posts $N_{c \mid d_i}$ to include in the sample:
$$N_{c \mid d_i} = P(c \mid d_i) \cdot N$$
4) *Rank posts by relevance:* For each class $c \in C'(d_i)$, rank all posts $p \in D$ by their probability $P(c \mid d_i)$ in descending order.
5) *Select top posts:* Select the top $N_{c \mid d_i}$ posts with the highest $P(c \mid d_i)$ values for each class $c$, forming a subset $S(c \mid d_i)$ of posts for that class.
6) *Combine subsets:* Aggregate the subsets across all selected classes and dimensions to form the final sample:
$$S = \bigcup_{d_i} \bigcup_{c \in C'(d_i)} S(c \mid d_i)$$

This method ensures that the sample $S$ is representative of the class distributions across the selected dimensions and classes, aligning with the objectives of the specific prompt. By focusing on relevant dimensions, such as topics, sentiment, emotions, or specific locations, the approach enables the creation of targeted and informative reports, providing stakeholders with valuable insights to navigate disaster events effectively.

Below, as done before, we present the main prompts used to generate the media report (topic and sentiment) and the reports for supporting emergency services (sub-events reported by users in a specific city involved in the disaster).

> `"topic_prompt"`: "Produce an analytical report summarizing the key topics discussed in posts by citizens living in the affected area (*$A*) during the disaster event (*$E*) within the date range (*$D*). The report has to be constrained to (*$W*) words, presented in paragraph format without subsections. Do not include a conclusion. Use as input the following social media posts: $[post_1, ..., post_N]$."
> `"opinion_prompt"`: "Generate an analytical report detailing the sentiments and emotions expressed by users regarding the disaster event (*$E*) in the affected area (*$A*) within the date range (*$D*). The report has to be constrained to (*$W*) words, presented in paragraph format without subsections. Exclude a conclusion section. Use as input the following social media posts: $[post_1, ..., post_N]$."
> `"city_subevent_prompt"`: "Produce an analytical report to support three key stakeholders (Police, EMS, and Firefighters) by summarizing actions taken, issues reported by citizens, and tasks to address in the affected location (*$C*) during the disaster event (*$E*) within the data range (*$D*). The report has to be constrained to (*$W*) words, presented in paragraph format without subsections. Use as input the following social media posts: $[post_1, ..., post_N]$. Insert in the report references to original posts using a bibliographic style (e.g., [1] $post_1$, [2] $post_2$, ...)."

### D. Case study: Camp Fire (2018)

In this section, we discuss how the two approaches described above, *basic* and *advanced*, use prompts to generate reports specifically tailored for media operators and emergency services. For brevity, we focus on example reports from a specific case study, the 2018 Camp Fire in Northern California, and provide a coverage analysis of the generated reports. The Camp Fire of 2018 is recognized as the deadliest and most destructive wildfire in California's history, claiming 85 lives, destroying over 18,000 structures, and causing an estimated $16.5 billion in damages.

*1) Report for media operators on topics of discussion:* First, we analyze the process of generating reports for media operators using the `topic_prompt` and how discussion topics are characterized. For clarity, we present two excerpts of reports generated for the 2018 Camp Fire (Figure 4), comparing the basic and advanced approaches. In the advanced approach, which utilizes a data sample, Figure 5 illustrates the 24 discussion topics extracted from posts about the event, represented as $T0$ to $T23$. Topics are identified using BERTopic and visualized as clusters compressed via UMAP. Such topics cover a wide range of themes, including the death toll and destruction caused by the fire, political debates about forest management, support efforts for victims, health and environmental concerns, and gratitude towards first responders. Star-shaped points indicate the approximately 100 tweets selected as input for the advanced prompt to generate the report.

The *basic approach* summarizes key issues reported by citizens, such as resource shortages, evacuation challenges, and infrastructure failures. While effective at highlighting immediate concerns, it provides a surface-level analysis focused on urgent problems and citizen responses. In contrast, the *advanced approach* offers a deeper analysis, addressing not only immediate impacts like destruction and displacement but also broader themes, such as political debates, health concerns,

> *[BASE]:* Many social media users emphasized the dire need for resources and assistance **[T2]**. Posts detailed missing persons and the devastating displacement of thousands of people, creating a sense of urgency for effective search and rescue operations **[T15]**. Others sought to amplify calls for help, appealing to influencers and organizations to step in and provide relief. This grassroots mobilization underscored the gaps in immediate aid distribution and communication channels during the crisis **[T2]**. The wildfire's rapid spread overwhelmed existing evacuation strategies. Several posts pointed to rethinking California's firefighting tactics and evacuation plans to ensure in preparedness in future issues **[T16]**.
>
> *[ADV]:* The fire caused unprecedented devastation **[T1]**, destroying entire towns such as Paradise **[T7]**, leaving thousands of homes **[T12]** and businesses reduced to ashes **[T10]**. Many posts focused on the heartbreaking scale of the destruction and the growing death toll, which climbed steadily as recovery efforts continued **[T4]**. Residents shared accounts of mandatory evacuations, with many fleeing their homes under dire conditions **[T16]**. Some pointed to the harrowing reality of makeshift camps in parking lots, where displaced families sought refuge **[T9]**. Efforts to provide temporary housing, including Airbnb's initiative to offer free stays for evacuees and emergency workers, were widely appreciated **[T2]**.

Fig. 4. Excerpts from the reports for media operators on topics of discussion during the 2018 Camp Fire: Basic vs. Advanced approaches.

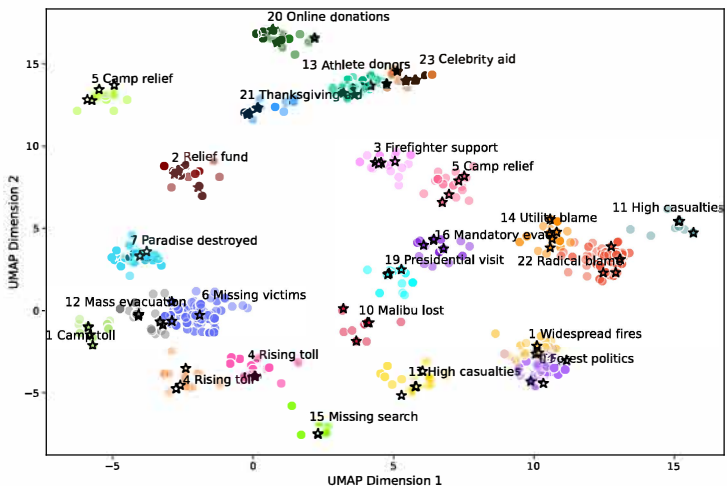

Fig. 5. Clustered representation of Twitter discussions during the 2018 Camp Fire, with topics extracted using BERTopic and compressed via UMAP. Star-shaped points indicate posts used as input for the advanced prompt to generate the report.

and public solidarity. By integrating diverse perspectives, the advanced method provides a more comprehensive view of the wildfire's societal and environmental implications. To numerically evaluate the coverage of discussion topics, we employed the *ChatGPT o1* model, selected for its advanced reasoning capabilities. The results generated by the *o1* model were carefully reviewed and manually validated to ensure accuracy. Even so, the model's use enables automation, facilitating more efficient and consistent analysis of various aspects, such as topics, opinions, and sub-events, across different case studies. Specifically, the model was tasked with assessing how effectively the 24 topics extracted by BERTopic for the Camp Fire use case, as visualized in Figure 5, are represented in the generated reports. To ensure unbiased results, the evaluation was conducted without disclosing whether the report was generated using the basic or advanced approach. This task was repeated ten times to verify consistency, with each iteration performed in a new temporary session that does not retain data or leverage information from prior analyses.

The basic approach achieved a topic coverage on average of 48%, whereas the advanced approach demonstrated a significant improvement, covering 87% of the topics. This enhanced performance is attributed to the advanced method's use of a filtered and targeted set of posts, allowing for the extraction of more precise and relevant information compared to the broader and less focused analysis of an unfiltered dataset.

*2) Report for media operators on user opinions:* Second, we examine how the `opinion_prompt` is used to generate reports for media operators and how discussion opinions (sentiment and emotions) are described. The basic approach processes a file containing the entire dataset, whereas the advanced approach utilizes a balanced sample of posts, constructed to ensure representation of both sentiments (positive and negative) and various emotions (e.g., anger, fear). Figure 6 compares excerpts from the basic and advanced reports on user opinions during the 2018 Camp Fire, illustrating that the advanced approach provides a more detailed and diverse analysis, integrating varied perspectives and concrete examples beyond the broad trends captured by the basic approach.

> *[BASE]:* The general sentiment revealed an overall negative trend, indicating widespread distress and concern **[sadness, fear]**. Subjectivity suggests that while many posts were emotionally charged **[anger, fear]**, others provided factual updates or broader reflections **[anticipation]**. Posts highlighting personal and communal tragedy dominated the narrative. For instance, one user lamented the loss of life and the missing persons during the fire **[sadness]**. Similarly, another post conveyed despair with the stark update, amplifying the emotional toll on both those directly and indirectly affected **[sadness, fear]**.
>
> *[ADV]:* Reports highlight the immense scale of the disaster, with descriptions of the Camp Fire as "the most destructive wildfires in California history" **[fear, sadness]**. Some posts highlight the deep collective mourning for lives and property lost, with posts calling for swift relief efforts to address the overwhelming needs of those affected **[sadness, anticipation]**. Posts celebrated acts of bravery, such as a bus driver saving children trapped in the flames **[joy, optimism]**. Political tensions flared, as some posts labeled the president's remarks as demeaning and lacking empathy for those suffering **[anger, sadness]**.

Fig. 6. Excerpts from the reports for media operators on user opinions during the 2018 Camp Fire: Basic vs. Advanced approaches.

We used *ChatGPT o1* to analyze a separate sample of 100 tweets, determining how many opinions from these tweets were addressed in the reports generated by each approach. The results show that the advanced report effectively captures nearly all situations represented in the reference sample. For instance, it encompasses a broad range of emotions, including positive sentiments (such as resilience, community support, heroism, and relief efforts) as well as negative sentiments (such as loss, anger, frustration, and political blame). Opinion coverage was significantly higher with the advanced approach, which addressed 96% of the opinions in the sample, compared to only 22% achieved by the basic approach. By focusing on a set of filtered and targeted posts, the advanced approach minimizes noise and dilution, allowing for the extraction of more precise and complete insights than the broader, less targeted analysis of an unfiltered dataset.

*3) Report for emergency services:* Third, we evaluate the reports generated for emergency services using the `city_subevent_prompt`, focusing on how sub-events occurring in a specific location (e.g., a city) were covered and described. Figure 7 provides an excerpt from the reports generated for the police regarding the city of Paradise, California, during the 2018 Camp Fire, comparing the basic and advanced approaches. The advanced report provides a detailed account of police efforts, emphasizing evacuation challenges, search and rescue operations, and looting prevention, while the basic report focuses on general safety concerns, highlighting traffic issues, looting fears, and missing persons. To measure the level of coverage, we tasked *ChatGPT o1* with verifying how many of the sub-events related to Paradise were included in the reports. The analysis considered 10 sub-events described in 418 posts (many posts referenced the same sub-event). The advanced approach achieved a coverage of approximately 80%, while the basic approach covered only 60% of the sub-events. This demonstrates the ability of the advanced approach to better capture and describe critical sub-events in the reports.

> **[BASE]:** During the Camp Fire, residents frequently reported concerns related to safety and evacuation routes. Numerous posts highlighted blocked roads and traffic jams impeding safe evacuations **[1, 2]**. Citizens also expressed fear of looting in abandoned homes and businesses, emphasizing the need for heightened security measures in evacuated areas **[3, 4]**. Missing persons were a recurring theme, with family members desperately seeking updates about their loved ones **[5, 6]**.
>
> **[ADV]:** Early on, evacuation notices were issued, but reports suggest that traffic gridlocks and poor planning led to many residents being trapped, exacerbating the death toll **[7, 8, 9]**. Law enforcement has also been tasked with searching for missing persons and identifying victims. The scale of the disaster left thousands unaccounted for, with over 600 people reported missing at its peak **[10, 11]**. Efforts have also included patrolling to prevent looting in the devastated town. Survivors have expressed frustration over the chaotic evacuation process, highlighting the need for improved disaster response planning **[12, 13]**.
>
> [1] "california battle, roads emerged major vulnerability escaping…"
> [2] "town escape routes turned traffic jams, fire popped tires melted…"
> …..

Fig. 7. Excerpts from the reports for the police on the city of Paradise, California, during the 2018 Camp Fire: Basic vs. Advanced approaches.

*4) Interactive report:* After introducing the reports generated for the police (with similar reports easily adaptable for other stakeholders), we now demonstrate how interactivity can further enhance the utility of these reports. The ability to interact dynamically with the generated content enables stakeholders, such as emergency services, to retrieve specific, actionable information efficiently, tailored to their immediate needs. Figure 8 shows an example of interactivity with the chatbot regarding a report generated for the police. The example demonstrates the chatbot's ability to answer targeted questions, such as the number of victims or destroyed houses in Paradise, California, during the Camp Fire. The chatbot provides detailed responses drawn directly from the report, offering both context and quantitative data, such as the destruction of approximately 13,972 homes, along with qualitative insights into the scale of the disaster. Additionally, it highlights scenarios where the police can assist other emergency services, such as EMS and firefighters, emphasizing the interconnected nature of disaster response efforts. This interactivity showcases the value of the chatbot as a dynamic tool for emergency services, allowing stakeholders to query and access relevant information quickly. By facilitating targeted inquiries, the service enhances situational awareness, supports decision-making, and improves coordination during critical operations.

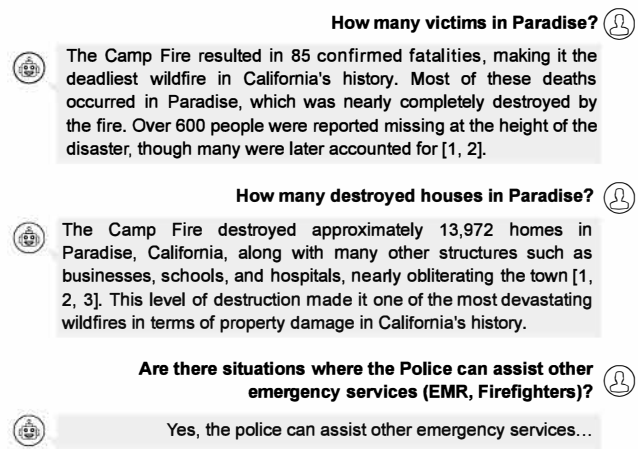

Fig. 8. Example of interactivity with the chatbot on a report generated for the police.

*E. Performance evaluation*

This section examines the performance of the two previously introduced approaches, *basic* and *advanced*, to demonstrate that, regardless of the disaster event, appropriately filtering data for ChatGPT using additional classes, carefully selected according to the prompt, enhances its ability to generate detailed and accurate reports. To this end, we generated reports for eight different disasters using both approaches, following the prompts described in the previous section. The quality of these reports was then evaluated using the following three methods:

*1) Textual Analysis and Quality Metrics:* This evaluation focuses on assessing the quality of the generated reports using both coverage-based and metric-based approaches. Specifically, the analysis includes coverage of key aspects such as topics discussed and opinions expressed in reports for media operators, as well as sub-events identified in reports for emergency services. In addition, we employ a range of textual analysis metrics to evaluate the report quality. These include frequency-based metrics like TF-IDF, N-gram overlap metrics (i.e., ROUGE-1, ROUGE-2, and ROUGE-L), semantic similarity measures using S-BERT and S-RoBERTa, and overall text similarity using BERTScore. Furthermore, comprehension-based metrics like BLANC-help and BLANC-tune are used to assess the report coherence and readability. These parameters are widely adopted for measuring summary quality in terms of relevance, coherence, and completeness compared to reference texts [46], [47].
2) *ChatGPT Evaluation*: in this process, *ChatGPT 4o* itself evaluates the report by giving scores based on criteria such as informativeness, quality, coherence, attributability, and overall impression [48].
3) *Domain Expert Evaluation*: it involves obtaining evaluations from experts who assess the reports and give them scores based on informative content, technical aspects, clarity of presentation, and precision. In such a way, the experts collectively choose the best report based on their expertise and judgment.

*1) Textual Analysis and Quality Metrics:* We first examine the extent to which the generated reports cover the key information shared by users during disaster events. Table II presents the percentage coverage of topics, opinions, and sub-events achieved by the basic and advanced approaches across eight natural disasters. Topic coverage was calculated by comparing the representation of topics extracted for each considered disaster in the generated reports. Similarly, opinion coverage was determined by analyzing a sample of one hundred tweets extracted from each case study and identifying the percentage of opinions addressed in the reports. Sub-event coverage was assessed by examining how accurately the reports described the sub-events in one of the major cities most affected by each disaster, in order to evaluate the completeness of the reports. The disasters analyzed and their corresponding most affected cities are listed below:

- *Hurricanes*: Hurricane Harvey (Houston) and Hurricane Irma (Miami).
- *Earthquakes*: Italy Earthquake (Amatrice) and Mexico Earthquake (Mexico City).
- *Floods*: Sri Lanka Flood (Colombo) and Kerala Flood (Kochi).
- *Wildfires*: Canada Wildfire (Fort McMurray) and California Wildfire (Paradise).

The results demonstrate that the advanced approach provides superior coverage compared to the basic approach across all key dimensions (topics, opinions, and sub-events), highlighting its ability to generate more comprehensive reports. Below, we discuss the coverage results for each dimension in detail:

- Topics: the advanced approach consistently outperforms the basic approach, achieving a significant increase (about 38% on average) in topic coverage for all events. For example, during Hurricane Irma (Miami), the advanced approach achieves 72% coverage, compared to 32% for the basic approach.
- Opinions: a similar trend is observed for opinion coverage, where the advanced approach significantly enhances the results (about 64% on average). As an example, during the Sri Lanka Flood (Colombo), the advanced approach achieves 97% coverage, compared to 35% for the basic approach.
- Sub-events: for disasters where sub-event information is available, the advanced approach also demonstrates superior performance, with an average improvement of about 28%. For instance, during the Canada Wildfire (Fort McMurray), the advanced approach achieves 88% coverage, compared to 82% for the basic approach.

These results clearly demonstrate that incorporating additional class information and appropriately filtering the data, as done in the advanced approach, significantly improves the reports' ability to capture critical details. By focusing on topics, opinions, and sub-events for the main affected cities, the advanced approach ensures a more comprehensive and accurate representation of the events compared to the basic method.

TABLE II
COVERAGE ANALYSIS OF TOPICS, OPINIONS, AND SUB-EVENTS (FOR THE MAIN AFFECTED CITY) DISCUSSED BY SOCIAL MEDIA USERS DURING THE DISASTER EVENTS USING BASIC AND ADVANCED APPROACHES.

|  | Coverage in reports (%) | | | | | |
|---|---|---|---|---|---|---|
|  | *Topics* | | *Opinions* | | *Sub-events* | |
|  | Basic | Adv | Basic | Adv | Basic | Adv |
| **Hurricane Harvey** | 0.36 | 0.64 | 0.19 | 0.92 | 0.21 | 0.46 |
| **Hurricane Irma** | 0.32 | 0.72 | 0.24 | 0.94 | 0.47 | 0.98 |
| **Earthquake Italy** | 0.67 | 0.87 | 0.41 | 0.95 | 0.71 | 0.79 |
| **Earthquake Mexico** | 0.60 | 0.87 | 0.38 | 0.88 | 0.30 | 0.50 |
| **Flood Sri Lanka** | 0.40 | 0.90 | 0.35 | 0.97 | 0.21 | 0.79 |
| **Flood Kerala** | 0.30 | 0.75 | 0.25 | 0.95 | 0.50 | 0.63 |
| **Wildfire Canada** | 0.32 | 0.84 | 0.38 | 0.96 | 0.82 | 0.88 |
| **Wildfire California** | 0.48 | 0.87 | 0.22 | 0.96 | 0.60 | 0.80 |

Afterward, we employ a range of textual analysis metrics to evaluate the quality of the generated reports. Table III presents the scores derived from the reports generated using social media data for the different disasters. To assess the quality of the reports against the original text, we utilized a set of widely adopted metrics. In the absence of a specific reference text, we defined the reference as the concatenated text of all posts describing the event. The following metrics were considered:

- *Lexical similarity (TF-IDF)*: it is calculated as the cosine similarity between TF-IDF vectors of the concatenated text and the report, after removing stop words and stemming. This metric quantifies similarity in word usage and distribution, providing insights into semantic correspondence [49].
- *N-gram overlap*: this includes metrics that evaluate the overlap between the report and the original text using unigrams (*Rouge-1*), bigrams (*Rouge-2*), and the longest common subsequence between the concatenated text and the report (*Rouge-L*) [50].
- *Semantic similarity*: it includes metrics that leverage cosine similarity between embeddings of the concatenated text and the report using *S-BERT* and *S-RoBERTa*. These metrics measure contextual similarity between texts.
- *Bert-Score*: this metric utilizes deep learning to evaluate similarity between sentences of the concatenated text and the report, focusing on word-level semantic relationships.
- *BLANC metrics*: these assess coherence and quality by analyzing additional information and necessary adjust-

ments to make the report understandable and accurate relative to the original text. They include two metrics, namely BLANC-help and BLANC-tune [51].

TABLE III
EVALUATION OF THE QUALITY OF REPORTS FOR MEDIA OPERATORS GENERATED BY BASIC AND ADVANCED APPROACHES USING DIVERSE METRICS.

| | Report quality (scores in %) | | | | | | | |
|---|---|---|---|---|---|---|---|---|
| | Hurricanes | | Earthquakes | | Floods | | Wildfires | |
| | Basic | Adv | Basic | Adv | Basic | Adv | Basic | Adv |
| **TF-IDF** | 0.40 | 0.49 | 0.44 | 0.49 | 0.37 | 0.43 | 0.38 | 0.46 |
| **Rouge-1** | 0.28 | 0.32 | 0.28 | 0.32 | 0.32 | 0.39 | 0.26 | 0.44 |
| **Rouge-2** | 0.21 | 0.38 | 0.32 | 0.39 | 0.32 | 0.36 | 0.30 | 0.37 |
| **Rouge-L** | 0.04 | 0.08 | 0.03 | 0.07 | 0.03 | 0.07 | 0.04 | 0.08 |
| **S-BERT** | 0.61 | 0.64 | 0.56 | 0.61 | 0.55 | 0.61 | 0.50 | 0.64 |
| **S-RoBERTa** | 0.63 | 0.66 | 0.63 | 0.72 | 0.60 | 0.64 | 0.56 | 0.60 |
| **Bert-Score** | 0.49 | 0.57 | 0.55 | 0.58 | 0.56 | 0.59 | 0.55 | 0.60 |
| **BLANC-help** | 0.18 | 0.28 | 0.18 | 0.24 | 0.23 | 0.24 | 0.21 | 0.23 |
| **BLANC-tune** | 0.34 | 0.42 | 0.32 | 0.43 | 0.33 | 0.32 | 0.32 | 0.38 |

Upon reviewing the scores, a consistent upward trend is observed across all metrics when transitioning from the basic to the advanced version of the approach. IN particular, TF-IDF and Rouge-L exhibit more pronounced improvements, indicating that the advanced version provides more comprehensive descriptions that better capture and synthesize the essence of user posts. Other metrics, such as Rouge-1, Rouge-2, S-BERT, S-RoBERTa, BERTScore, BLANC-help, and BLANC-tune, also show notable enhancements, which highlight the effectiveness of the advanced approach in enhancing summary quality across various evaluation dimensions.

*2) ChatGPT and Domain Expert Evaluation:* This qualitative assessment involves ChatGPT rating each report on a scale of 1 (worst) to 5 (best) across the five defined criteria.

1) *Informative*: the report encapsulates crucial details from the source, offering a precise and concise presentation.
2) *Quality*: the report is understandable and comprehensible, demonstrating high quality.
3) *Coherence*: the report exhibits a sound structure and organization, ensuring coherence.
4) *Attributable*: all information in the report is attributable to the source.
5) *Overall Preference*: the report succinctly, logically, and coherently conveys the primary ideas from the source.

Figure 9(a) demonstrates that the advanced version emerges as the most preferable choice, showcasing significant improvements across multiple evaluation criteria compared to the basic version. Notably, the advanced version achieved consistently higher scores in all five evaluation metrics, with the greatest improvements observed in informativeness and quality. Conversely, the basic version consistently exhibited lower scores across all evaluation metrics. These enhancements are due to the advanced model's ability to read selected posts that are strictly connected to the prompt on which to create the report, whereas the basic version must find this information within the entire file containing all the posts collected for the disaster event

We conducted an evaluation with twenty domain experts to validate the explanations generated by the two approaches using ten different reviews. For each test, we presented excerpts from two reports (basic and advanced) generated from the same task (e.g., media report) and dataset (e.g., Camp Fire). The experts were asked to identify which report excelled in specific aspects. Specifically, they were asked to answer the following questions: ($i$) which report do you believe offers greater overall information content? ($ii$) which report contains more technical or specialized aspects? ($iii$) which report provides a clearer presentation? ($iv$) which report demonstrates greater precision and clarity in its contents? ($v$) which report do you prefer for overall quality? To ensure unbiased responses, the order of presentation of the basic and advanced versions was varied across the questions. Additionally, control systems were implemented to counter any potential order effects and ensure fair evaluation of both versions.

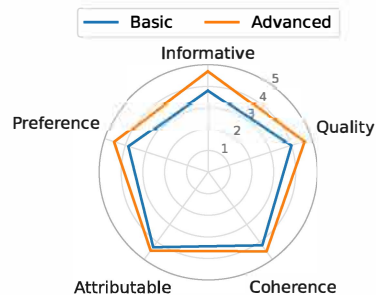

(a) ChatGPT evaluation.

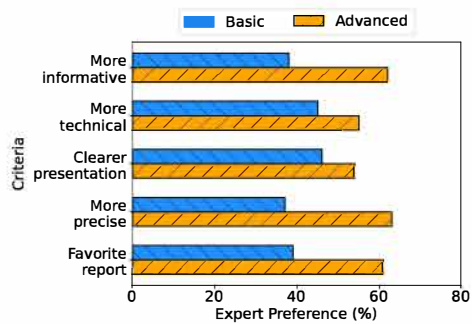

(b) Domain expert.

Fig. 9. Evaluation of reports for media operators generated by basic and advanced approaches, assessed by ChatGPT and domain experts.

Figure 9(b) illustrates the percentage of experts who preferred the basic and advanced versions across the five evaluation criteria. Domain experts consistently favored the advanced version over the basic version in all aspects. Notably, the advanced reports were rated significantly higher due to their greater informational content and precision in providing timely descriptions. Additionally, regarding technical aspects and clearer presentation, the writing capabilities of ChatGPT effectively amalgamate and link different contents, further reducing the gap between the basic and advanced versions. These findings underscore the superiority of the advanced approach in meeting expert standards across multiple evaluation criteria.

## V. Conclusion

In recent years, social media platforms have become indispensable tools for understanding human dynamics, offering vast amounts of real-time information during disasters, catastrophic events, and in shaping sustainable urban management. These platforms provide critical insights into urban challenges, including resource allocation, transportation systems, and environmental monitoring, thereby supporting the development of smarter and more resilient cities. Despite advancements in machine learning techniques for classifying and aggregating social media content, a critical need remains to further enhance the automation, aggregation, and organization of citizen-reported issues. Our proposed methodology bridges this gap by leveraging the capabilities of full-spectrum Large Language Models (LLMs) to comprehensively analyze and synthesize user-generated content from disaster-affected areas. By combining BERT models for precise, multidimensional classification with generative models such as GPT-4 to produce detailed summaries and customized reports, our approach ensures the creation of actionable insights tailored to diverse stakeholders, including emergency services, press outlets, and operational teams.

Extensive experiments on diverse datasets validate the efficacy of our methodology in detecting events and issues during and after disasters. By preliminary enriching posts by classifying them on multiple dimensions, such as type, location, sentiment, and topics, we enable generative AI tools to produce more precise and contextually relevant reports. These reports are further tailored to meet the unique needs of various stakeholders through interactive features that facilitate data exploration, refine outputs, and customize report formats. For example, EMS teams may request concise, location-specific summaries, while others might prefer narrative overviews or structured data tables. This adaptability ensures that stakeholders receive the most relevant and actionable information.

Quantitative evaluations using text scores and latent representations, alongside qualitative assessments from automated tools and field experts, highlight the superiority of our approach over traditional methods, including baseline systems like ChatGPT. In particular, quantitative results show the advanced approach achieves a 38% increase in topic coverage, a 64% improvement in opinion coverage, and a 28% boost in sub-event coverage.

Future work will focus on further refining the scalability, adaptability, and robustness of our methodology across various disaster scenarios and geographic regions. Enhancing real-time processing capabilities will remain a priority to ensure timely dissemination of critical information. Moreover, we aim to integrate more advanced features into our LLM-powered system, enabling it to handle the diverse and dynamic nature of social media content with greater efficiency. By continually innovating and expanding the system's capabilities, we seek to advance disaster preparedness, response, and resilience, ultimately contributing to more effective management and mitigation efforts.

## Acknowledgment

This work was supported by the research project "INSIDER: INtelligent ServIce Deployment for advanced cloud-Edge integRation" granted by the Italian Ministry of University and Research (MUR) within the PRIN 2022 program and European Union - Next Generation EU (grant n. 2022WWSCRR, CUP H53D23003670006) and by the European Union under the Italian National Recovery and Resilience Plan (NRRP) of NextGenerationEU, partnership on "Telecommunications of the Future" (PE00000001 - program "RESTART"). We also acknowledge financial support from "National Centre for HPC, Big Data and Quantum Computing", CN00000013 - CUP H23C22000360005